\documentclass[conference]{IEEEtran}
\IEEEoverridecommandlockouts

\usepackage{graphicx}
\usepackage{booktabs}
\usepackage{amsmath}
\usepackage{amssymb}
\usepackage{xcolor}
\usepackage{balance}
\usepackage{enumitem}
\usepackage{subcaption}
\usepackage{multirow}
\usepackage{tabularx}
\usepackage{microtype}

\usepackage{tcolorbox}

\newtcolorbox{qbox}{
  colback=gray!10,
  colframe=gray!50,
  boxrule=0.5pt,
  arc=2pt,
  left=5pt,
  right=5pt,
  top=3pt,
  bottom=3pt
}

\begin{document}

\title{Jiao: Bridging Isolation and Customization in Mixed Criticality Robotics\\
\thanks{$^{*}$These authors contributed equally to this work.}
}

\author{
\IEEEauthorblockN{
James Yen\textsuperscript{*1}, Zhibai Huang\textsuperscript{*1}, Zhixiang Wei\textsuperscript{1}, Tinghao Yi\textsuperscript{2}, Shupeng Zeng\textsuperscript{2},\\
Liang Pang\textsuperscript{2}, Songtao Xue\textsuperscript{1}, Zhengwei Qi\textsuperscript{1}
}
\IEEEauthorblockA{
\textsuperscript{1}\textit{Shanghai Jiao Tong University}, Shanghai, China\hspace{0.8em}
\textsuperscript{2}\textit{OpenMind}, China
}
\IEEEauthorblockA{
\scriptsize
\{jamesyen2202002,paynqueller,tonywei\_sjtu,xxxlhhxz,qizhwei\}@sjtu.edu.cn,
\{yitinghao,zengshupeng,pangliang\}@efort.com.cn
}
}

\maketitle

\begin{abstract}
Consumer robotics demands consolidation of safety-critical control, perception pipelines, and user applications on shared multicore platforms. While static partitioning hypervisors provide hardware-enforced isolation, directly transplanting automotive architectures encounters an expertise asymmetry problem in which end-users modifying robot behavior lack the systems knowledge that platform developers possess. We present an architecture addressing this challenge through three integrated components. A Safe IO Cell provides hardware-level override capability. A Parameter Synchronization Service encapsulates cross-domain complexity. A Safety Communication Layer implements IEC~61508-aligned verification. Our empirical evaluation on an ARM Cortex-A55 platform demonstrates that partition isolation reduces cycle-period jitter by 84.5\% and cuts tail timing error by nearly an order of magnitude (p99 $|$jitter$|$ from 69.0\,$\mu$s to 7.8\,$\mu$s), eliminating all $>$50\,$\mu$s~excursions.
\end{abstract}

\section{Introduction}

Industrial robots have evolved from dedicated controllers into sensor-rich computing platforms that integrate multi-camera perception, LiDAR-based SLAM, learned models, and continuous software updates~\cite{macenski2022robot,cadena2016slam}. Cost, weight, and power constraints drive teams to consolidate these workloads onto fewer system-on-chip (SoC) platforms~\cite{anderson2006real}, making contention on shared CPUs, caches, memory, and accelerators routine. This threatens control latency and erodes safety margins. Safety-critical industries faced similar pressures and adopted static partitioning, where hard real-time control executes in protected cells while best-effort software runs elsewhere, as standardized by AUTOSAR~\cite{furst2016autosar} and ARINC~653~\cite{arinc653p0_4}.

\begin{figure}[t]
    \centering
    \includegraphics[width=\columnwidth]{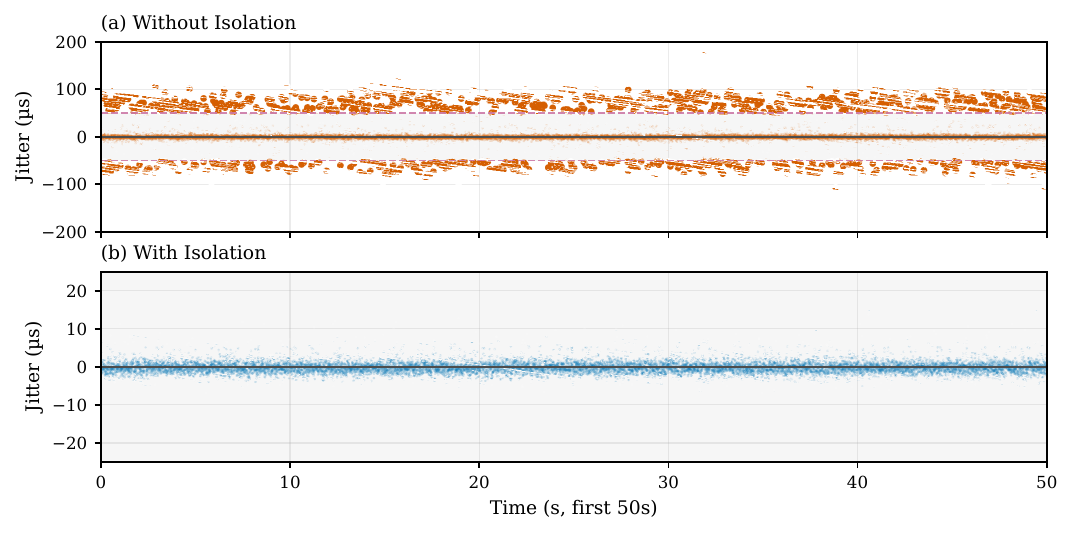}
    \caption{Control-cycle timing jitter on a 6-DOF manipulator. (a)~Baseline with concurrent perception workloads exhibits heavy-tailed jitter. (b)~Partition isolation stabilizes timing and eliminates large excursions.}
    \label{fig:jitter-motivation}
\end{figure}

Lacking a standardized isolation model, robotics teams have historically achieved separation through physical decomposition, dedicating microcontrollers to servo control and separate Linux computers to perception. This workaround fails to scale as robots incorporate additional sensors, accelerators, and connectivity, compounding cost, wiring complexity, and potential failure points. Consolidation onto shared hardware thus becomes inevitable, and the critical question is how to preserve the isolation guarantees that physical separation once~provided.

A natural first attempt is to make the consolidated system ``more real-time'' by reducing scheduling and interrupt latency. PREEMPT\_RT makes Linux fully preemptible but provides no spatial isolation. Dual-kernel architectures such as Xenomai~\cite{gerum2004xenomai} achieve low interrupt latency, yet processes still share a common address space. These mechanisms improve temporal scheduling but cannot prevent interference when perception and control contend for cache capacity, memory bandwidth, and thermal headroom. Virtualization-based partitioning addresses this by giving the control stack exclusive ownership of cores, memory, and dedicated~devices.

Figure~\ref{fig:jitter-motivation} illustrates the impact on a representative industrial manipulator. We estimate the 1\,kHz cycle period as $\Delta t_2{=}t[n]{-}t[n{-}2]$ and define jitter as deviation from the median. Under baseline conditions with concurrent perception workloads, cycle-period jitter has standard deviation 12.6\,$\mu$s with rare excursions exceeding 300\,$\mu$s. Under partition isolation, variance drops by 84.5\% and worst-case excursions shrink by nearly an order of magnitude, eliminating all excursions above~50\,$\mu$s.

\smallskip
\noindent \textbf{Challenge.}
Transplanting partitioning architectures into industrial robotics encounters the \textit{expertise asymmetry problem}. Automotive ECU configurations are fixed at manufacturing by engineers with complete knowledge of real-time scheduling. Industrial robots increasingly support runtime adaptation through teach pendants, vision-guided calibration, and cloud-connected parameter tuning. System integrators modify safety-relevant behavior without platform-level expertise. Bridging this asymmetry requires cross-partition interaction that is safe by construction, with abstractions that encapsulate protocol complexity and enforce validated parameter~application.

\smallskip
\noindent \textbf{Insight.}
Static partitioning, safety communication protocols, and hardware watchdogs are individually well established in automotive and industrial automation. However, robotics presents a distinct challenge. The partition boundary must simultaneously enforce safety integrity \textit{and} remain accessible to non-expert integrators who routinely modify safety-relevant parameters after deployment. Neither automotive ECU architectures (which assume fixed configurations) nor standalone fieldbus safety protocols (which assume trusted endpoints) address this combination. A complete solution must co-design hypervisor configuration, communication protocols, and hardware-level overrides into a unified safety architecture that encapsulates cross-domain complexity from the~integrator.

\smallskip
\noindent \textbf{Our Solution.}
We present \textsc{Jiao}, an architecture built on the Jailhouse static partitioning hypervisor that enables industrial robots to run hard real-time servo control, soft real-time perception, and operator applications on shared hardware while maintaining strict isolation. \textsc{Jiao} does not propose a new hypervisor. Instead, it treats the partition boundary as a safety interface enabling integrator-friendly customization. The system targets collaborative manipulators, mobile platforms combining SLAM with dynamic obstacle avoidance, and multi-axis assembly systems coordinating force control with learned insertion~policies.

\smallskip
\noindent \textbf{Key Contributions.}
\begin{itemize}[leftmargin=*,topsep=2pt,itemsep=1pt]
\item We design a \textbf{Safe IO Cell} serving as an independent safety authority on a dedicated partition, supervising heartbeats and command envelopes while maintaining hardware-level override capability for motor enables, brake engagement, and emergency stop~circuits.

\item We develop a \textbf{Parameter Synchronization Service} that lets integrators modify robot behavior through familiar ROS-style APIs without handling cross-partition protocols, shared memory layouts, or real-time scheduling~details.

\item We implement a \textbf{Safety Communication Layer} with IEC~61508-aligned integrity verification including cyclic redundancy checks, monotonic sequence numbering, and timestamp-based freshness validation, with progressive degradation semantics under partial communication~failure.
\end{itemize}

Beyond jitter reduction, the primary contribution is an architecture that governs cross-partition customization safely, reducing cycle-period jitter by 84.5\% and tail timing error by nearly an order of~magnitude.

\section{Background}
\label{sec:background}

\subsection{Mixed-Criticality Systems}

Mixed-criticality systems consolidate workloads with varying timing and safety integrity requirements onto a single platform~\cite{gheraibia2018overview}. Standards formalize these through integrity levels such as ASIL (ISO~26262~\cite{iso26262_1_2018}) and SIL (IEC~61508~\cite{iec61508_1_2010}), with freedom from interference as a fundamental requirement. Robotic architectures exemplify this tension. Motor control demands hard real-time execution, perception runs on soft real-time schedules, and user interfaces are best-effort. Conventional ROS deployments execute all three within shared memory spaces, complicating safety~certification.

\subsection{Static Partitioning Hypervisors}

Static partitioning hypervisors fix resource allocations at initialization, giving each partition dedicated CPU cores, memory regions, and device assignments~\cite{sinitsyn2015jailhouse,martins2020bao}. Unlike dynamic hypervisors that multiplex resources through complex runtime scheduling, this approach removes unpredictable policy decisions from the critical path. Jailhouse exemplifies this approach by activating after Linux boots, enforcing isolation through second-stage page tables and interrupt remapping, and providing inter-partition shared memory with doorbell interrupts. Communication semantics remain the responsibility of partition~software.

\begin{figure}[t]
    \centering
    \includegraphics[width=\columnwidth]{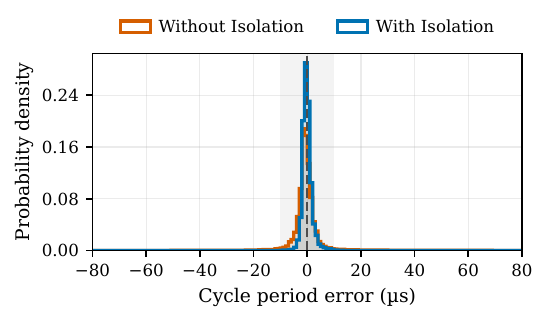}
    \caption{Distribution of estimated 1\,kHz control-cycle periods. Isolation substantially tightens the distribution and reduces tail~excursions.}
    \label{fig:cycle-distribution}
\end{figure}

\subsection{Safety Communication Requirements}

IEC~61508 specifies the ``black channel'' principle for safety-relevant communication over untrusted media~\cite{iec61508_1_2010}. Under this model, the transport may corrupt, lose, delay, duplicate, or reorder data, so safety derives entirely from endpoint protocol mechanisms. Static partitioning ensures spatial isolation but does not validate data integrity, requiring a dedicated safety protocol over hypervisor shared~memory.

\section{Motivation}
\label{sec:motivation}

We studied representative mixed-criticality robotic deployments and derive three observations motivating our~design.

\subsection{Observations}
\label{sec:observations}

\noindent \textbf{Observation 1: System load fluctuations compromise real-time guarantees without isolation.}
ROS provides no fault containment between nodes of differing criticality, and corruption in perception pipelines can propagate to motor control. Figure~\ref{fig:cycle-distribution} quantifies this effect. Under baseline conditions the distribution is heavy-tailed (max~1.32\,ms), while under isolation it tightens substantially (max~1.03\,ms) and cycles within 1\,ms$\pm$10\,$\mu$s rise from 91.8\% to~99.7\%.

\begin{table}[t]
\caption{Safety criticality vs.\ customization frequency per ISO~10218-2~\cite{iso10218_2_2025}. Bold rows mark the critical risk~intersection.}
\label{tab:iso-analysis}
\centering
\small
\resizebox{\columnwidth}{!}{%
\begin{tabular}{lccc}
\toprule
\textbf{Component Class} & \textbf{Safety Impact} & \textbf{Tuning Freq.} & \textbf{Risk Level} \\ 
\midrule
Inverse Kinematics & High & Low & Low \\
Path Planning & Medium & High & Medium \\
\textbf{Servo Gains (PID)} & \textbf{High} & \textbf{High} & \textbf{Critical} \\
\textbf{Force Limits} & \textbf{High} & \textbf{High} & \textbf{Critical} \\
Vision/Inference & Low & High & Low \\
Fieldbus Driver & High & Low & Low \\
\bottomrule
\end{tabular}%
}
\vspace{-10pt}
\end{table}

\noindent \textbf{Observation 2: Post-deployment customization invalidates static safety assumptions.}
Consumer robots increasingly support modification through companion applications and scripting interfaces. Our analysis reveals a critical risk intersection where safety-critical components coincide with high customization frequency (Table~\ref{tab:iso-analysis}). Servo gains and force limits fall squarely in this zone, and a manufacturing engineer integrating proprietary algorithms may inadvertently modify these parameters without understanding implications for real-time~guarantees.

\begin{figure}[t]
    \centering
    \includegraphics[width=\columnwidth]{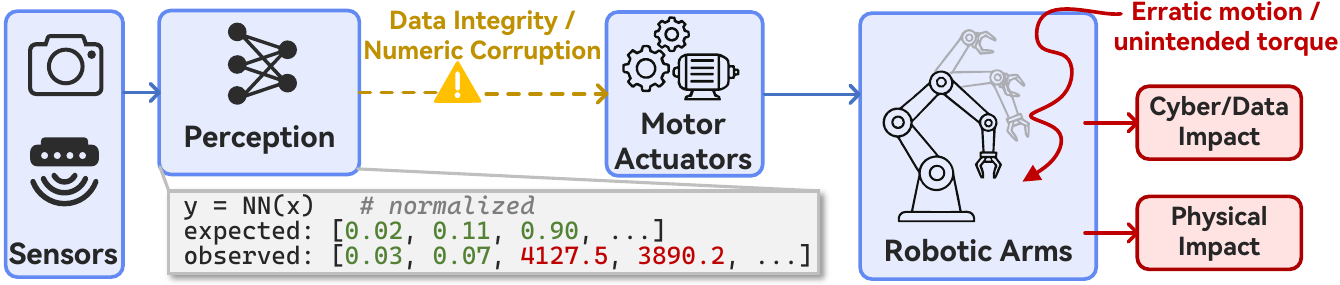}
    \caption{Failure propagation from neural-network inference corruption to physical actuator hazard. Out-of-range perception outputs bypass validation and drive motors into unsafe operating~regimes.}
    \label{fig:data-corruption}
\end{figure}

\noindent \textbf{Observation 3: Cross-domain data integrity failures cascade into physical hazards.}
When parameters traverse the non-real-time to real-time boundary, multiple failure modes can corrupt the control loop. Figure~\ref{fig:data-corruption} illustrates how numeric corruption in a perception network produces out-of-range outputs that propagate to motor actuators. Bit errors pass undetected without integrity verification, and message loss causes discontinuous state transitions. IEC~61508~\cite{iec61508_1_2010} mandates protocol-level verification that static partitioning hypervisors leave entirely to application~software.

\smallskip
\noindent \textbf{Key Insight.}
Static partitioning provides the foundation, but the isolation boundary requires active management through communication protocols that verify integrity, abstraction layers that hide complexity from non-expert users, and an independent safety authority with hardware-level override~capability.

\begin{figure*}[t]
    \centering
    \includegraphics[width=\linewidth]{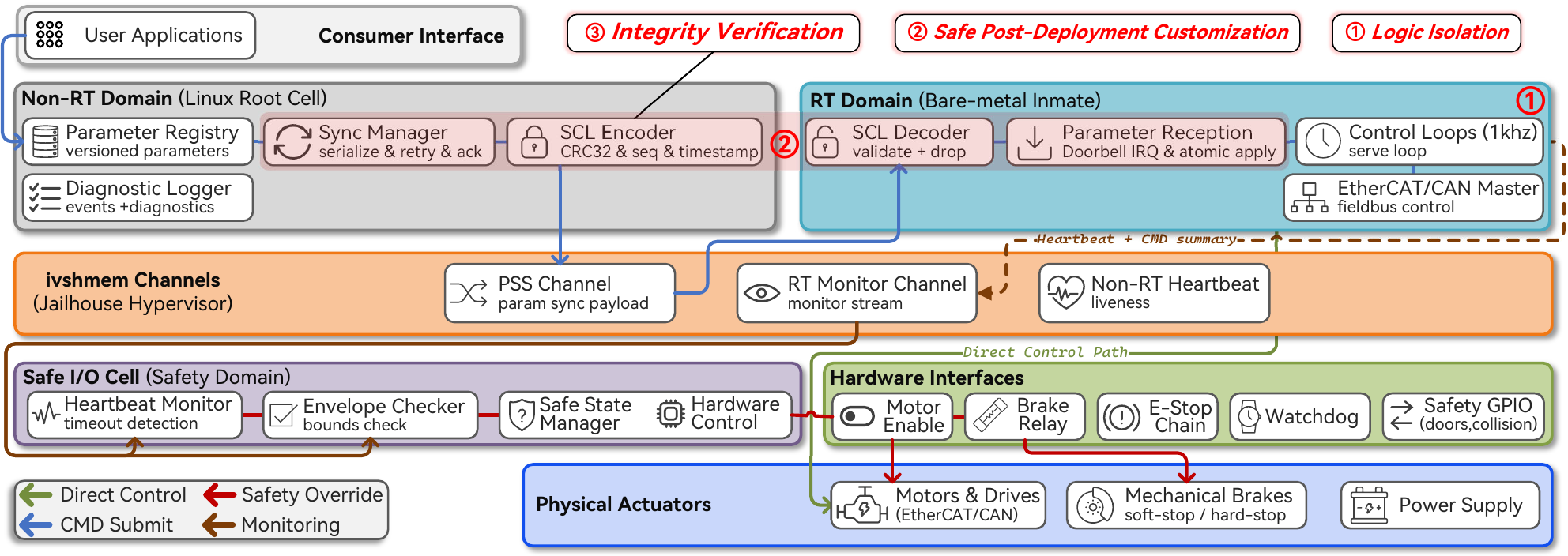}
    \caption{\textbf{System overview and contribution map.} A static partitioning hypervisor isolates the Non-RT Linux root cell, RT control cell, and Safe IO safety cell while enabling deterministic ivshmem communication. Our contributions span (\textcircled{1})~mixed-criticality separation with direct EtherCAT and CAN control paths, (\textcircled{2})~a parameter update pipeline via PSS, and (\textcircled{3})~an IEC~61508-aligned SCL layer that validates cross-domain data and drives Safe IO override through motor enable and brake~relay.}
    \label{fig:overview}
\end{figure*}

\subsection{Challenges}
\label{sec:challenges}

\begin{qbox}
\textbf{Q\#1.} How can we expose configuration interfaces to non-expert users without compromising real-time deadline guarantees or cross-domain synchronization~invariants?
\end{qbox}

\noindent\textbf{Challenge 1: Bridging Expertise Asymmetry.}
Platform developers understand inter-partition protocols but cannot anticipate deployment needs. End-users possess domain expertise but lack systems knowledge. As Table~\ref{tab:iso-analysis} shows, the most frequently tuned components carry the highest safety impact, demanding abstractions that enforce validated atomic parameter application without exposing cross-domain~complexity.

\begin{qbox}
\textbf{Q\#2.} How can we verify data integrity across the isolation boundary with guarantees sufficient for safety-critical real-time~control?
\end{qbox}

\noindent\textbf{Challenge 2: Guaranteeing Cross-Domain Integrity.}
Static partitioning provides spatial isolation but no integrity verification for shared-memory communication. Corrupted values crossing the partition boundary propagate unchecked to physical actuators (Figure~\ref{fig:data-corruption}), requiring IEC~61508-aligned verification with progressive degradation~semantics.

\begin{qbox}
\textbf{Q\#3.} How can we guarantee transition to a safe state when software-mediated control paths~fail?
\end{qbox}

\noindent\textbf{Challenge 3: Providing Hardware-Level Override.}
Software isolation cannot protect against unresponsive control partitions caused by hardware faults, infinite loops, or resource exhaustion. An independent safety authority with direct hardware access must force actuators into safe states regardless of primary controller~behavior.

\subsection{Design Goals}
\label{sec:design-goals}

We derive three goals addressing these~challenges.

\begin{itemize}[leftmargin=*,topsep=2pt,itemsep=1pt]
    \item \textbf{For Q\#1: Consumer-Transparent Configuration.} Encapsulate inter-partition complexity behind a ROS-compatible API that enforces validated atomic parameter application and mediates access to critical-risk components~(Table~\ref{tab:iso-analysis}).

    \item \textbf{For Q\#2: Safety-Aligned Communication.} Provide CRC, sequence numbering, and timestamp-based freshness validation with progressive degradation under partial communication~failure.

    \item \textbf{For Q\#3: Independent Safety Authority.} Place a Safe IO Cell on a dedicated partition with direct hardware control of motor enables, brakes, and emergency stops, supervising heartbeats and command envelopes independently of primary controller~software.
\end{itemize}

\section{Design}
\label{sec:design}

Figure~\ref{fig:overview} presents \textsc{Jiao}. The design extends static partitioning with three subsystems that treat the isolation boundary as a safety~interface.

\subsection{System Overview}

The system comprises three domains. The Non-RT Domain runs a Linux root cell with ROS~2 and user applications. The RT Domain executes deterministic control with direct fieldbus access. The Safe IO Cell occupies a dedicated safety partition with hardware control over actuator enables and emergency stop~circuits.

Communication uses ivshmem shared memory with two logical channels. The PSS Channel carries parameter updates to the RT Domain, and the RT Monitor Channel carries heartbeats and diagnostics in the reverse direction. The Safe IO Cell observes both channels independently and can override actuation without depending on the primary control~path.

\begin{figure}[t]
    \centering
    \includegraphics[width=\columnwidth]{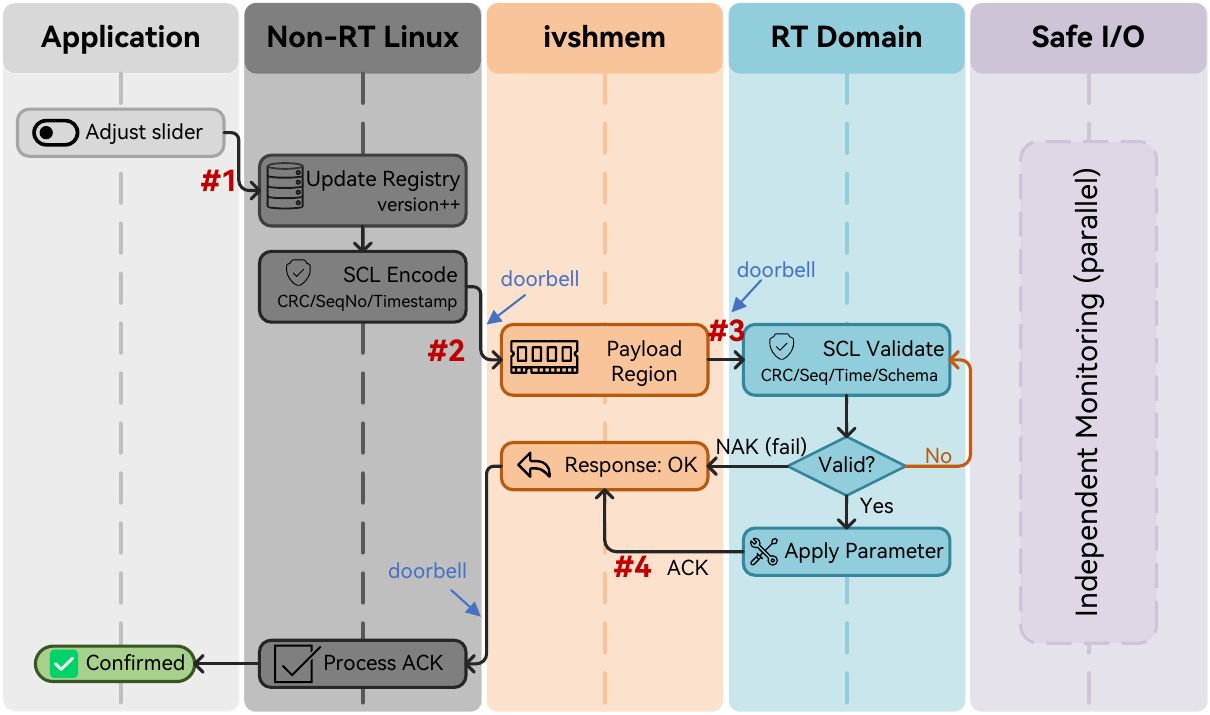}
    \caption{Parameter synchronization across Consumer Application, Non-RT Linux, ivshmem, and RT Domain. The Safe IO Cell monitors transactions and checks safety~envelopes.}
    \label{fig:cmd_flow}
\end{figure}

\subsection{Safe IO Cell}

The Safe IO Cell is an independent safety monitor with direct access to actuation safety interfaces. It owns GPIO lines controlling motor enable circuits, brake relays, and emergency stop interfaces. Deasserting motor enable or engaging brakes forces actuators into fail-safe configurations regardless of EtherCAT or CAN~commands.

The cell monitors heartbeats with configurable timeouts and checks commanded values against predefined envelopes. Its state machine governs transitions among normal operation, degraded modes, and safe states. Recovery requires explicit operator acknowledgment to prevent automatic resumption after transient~faults.

\subsection{Parameter Synchronization Service}

The Parameter Synchronization Service hides cross-domain mechanisms behind ROS-style parameter interfaces. Applications declare parameter types and bounds through standard APIs. The service validates updates, protects them with SCL, and applies them atomically inside the RT~Domain.

Figure~\ref{fig:cmd_flow} shows the update path. The Sync Manager packages an update with metadata and passes it to the SCL Encoder, which writes the protected message to the PSS Channel. A doorbell interrupt notifies the RT Domain, which validates and either applies or rejects the update, returning an acknowledgment. On failure, the RT Domain continues with last-known-good parameters while the Non-RT side~retries.

\subsection{Safety Communication Layer}

The Safety Communication Layer implements IEC~61508-aligned endpoint verification for all cross-domain traffic, treating shared memory as a black channel~\cite{iec61508_1_2010}. Each message carries a CRC-32C, a monotonic sequence number, and a timestamp. Table~\ref{tab:scl-coverage} summarizes failure-mode coverage. The SCL defines progressive degradation. Transient anomalies raise warnings while sustained failures drive transitions to safe stop through intermediate restricted states. Single-channel configurations target diagnostic coverage consistent with common SIL~2-oriented design patterns. Higher integrity levels can add redundant channels and replicated Safe IO units with~voting.

\begin{table}[t]
\centering
\caption{SCL failure-mode coverage following IEC~61508 black-channel design~patterns}
\label{tab:scl-coverage}
\small
\setlength{\tabcolsep}{3.5pt}
\begin{tabular}{@{}llll@{}}
\toprule
\textbf{Failure Mode} & \textbf{Mechanism} & \textbf{Transient} & \textbf{Persistent} \\
\midrule
Corruption & CRC-32C & Drop + retry & Safe stop \\
Replay/dup./reord. & Seqno window & Drop & Degrade \\
Delay/stale & Timestamp & Degrade & Safe stop \\
Loss & Timeout/HB & Retry & Safe IO override \\
\bottomrule
\end{tabular}
\end{table}

\section{Evaluation}
\label{sec:evaluation}

We evaluate timing isolation through empirical EtherCAT control-cycle measurements, comparing baseline against partition-isolated~configurations.

\subsection{Experimental Setup}

\noindent\textbf{Hardware Platform.}
We conduct experiments on the D9340 SoC (Table~\ref{tab:testbed}), a production-representative heterogeneous processor featuring a quad-core ARM Cortex-A55 cluster alongside a dedicated real-time~R-core.

\begin{table}[t]
\centering
\caption{Experimental testbed hardware configuration}
\label{tab:testbed}
\small
\renewcommand{\arraystretch}{1.08}
\setlength{\tabcolsep}{6pt}
\begin{tabular}{@{}ll@{}}
\toprule
\textbf{Component} & \textbf{Specification} \\
\midrule
SoC & D9340 Heterogeneous Processor \\
Application Cores & 4$\times$ ARM Cortex-A55 \\
Real-Time Core & Dedicated R-Core \\
Memory & 4\,GB LPDDR4 @ 3200\,MHz \\
Storage & eMMC Flash \\
Fieldbus & EtherCAT (IEEE 802.3br mPackets) \\
Host OS & Linux \\
\bottomrule
\end{tabular}
\end{table}

\noindent\textbf{Measurement Methodology.}
We capture EtherCAT traffic with nanosecond-resolution timestamps over 10-minute windows ($\approx$1.2M frames per trace). The \textit{baseline} trace captures system behavior under production-level mixed workloads from OpenMind~\cite{openmind_robotics}. The \textit{isolated} trace captures behavior with partition separation applied, replaying the same workload~traces.

\noindent\textbf{Metrics.}
To avoid mixing intra-cycle and inter-cycle gaps, we estimate the control-cycle period as $\Delta t_2{=}t[n]{-}t[n{-}2]$. We define the nominal period as $\mathrm{median}(\Delta t_2)$ and signed jitter as $\Delta t_2{-}\mathrm{median}(\Delta t_2)$. We report standard deviation~($\sigma$), tail percentiles (p99, p99.9), maximum absolute jitter, and \textit{large excursions} with $|$jitter$|{>}50\,\mu$s.

\subsection{Timing Isolation Effectiveness}

\begin{table}[t]
\centering
\caption{Control-cycle timing error derived from EtherCAT timestamps ($\Delta t_2{=}t[n]{-}t[n{-}2]$)}
\label{tab:timing-results}
\small
\setlength{\tabcolsep}{5pt}
\begin{tabularx}{\linewidth}{@{}Xrrr@{}}
\toprule
\textbf{Metric} & \textbf{Baseline} & \textbf{Isolated} & \textbf{Factor} \\
\midrule
Nominal cycle ($\mu$s)                 & 999.25 & 999.75 & -- \\
Jitter SD $\sigma$ ($\mu$s)            & 12.58  & 1.95   & \textbf{6.5$\times$} \\
p99 $|$jitter$|$ ($\mu$s)              & 69.0   & 7.8    & \textbf{8.8$\times$} \\
p99.9 $|$jitter$|$ ($\mu$s)            & 101.0  & 13.2   & \textbf{7.7$\times$} \\
Max $|$jitter$|$ ($\mu$s)              & 321.5  & 32.8   & \textbf{9.8$\times$} \\
$|$jitter$|{>}50\,\mu$s (\%)          & 2.04\% & 0.00\% & \textbf{--} \\
Missed cycles ($\Delta t_2{>}2$\,ms)   & 0      & 0      & -- \\
\bottomrule
\end{tabularx}
\end{table}

Table~\ref{tab:timing-results} summarizes control-cycle timing error. The baseline exhibits substantial cycle-period dispersion and heavy-tailed jitter, while the isolated configuration sharply reduces both typical jitter and tail~excursions.

\noindent\textbf{Jitter Reduction.}
Isolation reduces jitter standard deviation from 12.58\,$\mu$s to 1.95\,$\mu$s (\textbf{84.5\% reduction}), improving controller phase stability and reducing timing-induced disturbances in synchronized multi-axis~systems.

\begin{figure}[t]
    \centering
    \includegraphics[width=\columnwidth]{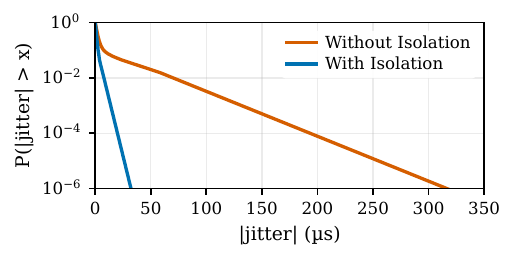}
    \caption{Tail distribution (CCDF) of absolute cycle-period jitter. Isolation reduces both typical and worst-case timing error by nearly an order of~magnitude.}
    \label{fig:cdf-tail}
\end{figure}

\noindent\textbf{Tail Risk Elimination.}
Tail timing error improves by nearly an order of magnitude. The p99~$|$jitter$|$ drops from 69.0\,$\mu$s to 7.8\,$\mu$s and worst-case excursion drops from 321.5\,$\mu$s to 32.8\,$\mu$s. The baseline contains 24,442 large excursions (2.04\% of cycles), while the isolated configuration exhibits zero. Figure~\ref{fig:cdf-tail} plots the complementary CDF of absolute cycle-period~jitter.

\begin{figure}[t]
    \centering
    \includegraphics[width=\columnwidth]{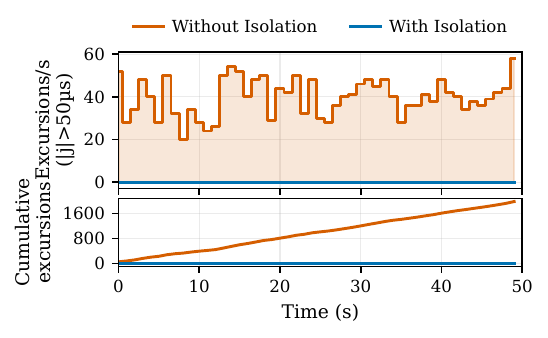}
    \caption{Large jitter excursions ($|$jitter$|{>}50\,\mu$s) per second over a 50-second window. Isolation eliminates all excursions above this~threshold.}
    \label{fig:deadline-misses}
\end{figure}

\noindent\textbf{Temporal Stability.}
As Figure~\ref{fig:deadline-misses} shows, the baseline exhibits 20--58 large excursions per second (mean~39.4) above the 50\,$\mu$s threshold, while the isolated configuration remains at zero throughout. The coefficient of variation improves from 1.26\% to 0.20\%, a 6.4$\times$ improvement in timing precision critical for coordinated multi-axis motion~control.

\smallskip
\noindent\textbf{Limitations.}
Our evaluation focuses on timing isolation under a single workload class (1\,kHz EtherCAT manipulator control). The contention sources we mitigate (shared CPU, cache, memory, and DMA) also arise in SLAM and vision-dominated pipelines~\cite{pellizzoni2010memory}, so partitioning benefits are expected to generalize. Quantifying black-channel fault detection via systematic fault injection remains important future~work.

\section{Related Work}

\noindent\textbf{Mixed-Criticality Theory.}
Vestal~\cite{vestal2007mc} formalized mixed-criticality scheduling, and subsequent work bounded memory interference on multicore platforms~\cite{pellizzoni2010memory}. These frameworks assume a priori task knowledge. \textsc{Jiao} targets deployments where integrators modify workloads post-deployment, requiring runtime-validated parameter application rather than offline analysis~alone.

\noindent\textbf{Real-Time Linux.}
PREEMPT\_RT~\cite{reghenzani2019preempt_rt} and Xenomai~\cite{gerum2004xenomai} improve scheduling latency but provide no spatial isolation. Priority-based ROS~2 executor scheduling~\cite{al2024dynamic} reduces average-case latency within a single OS instance. Our work adds hardware-enforced boundaries that kernel-level approaches cannot~provide.

\noindent\textbf{Static Partitioning Hypervisors.}
Jailhouse~\cite{sinitsyn2015jailhouse} partitions hardware resources after Linux boot. Bao~\cite{martins2020bao} targets minimal trusted computing base for embedded systems. ACRN~\cite{li2019acrn} provides partitioning for industrial IoT with safety certification goals. All three enforce spatial isolation but leave communication semantics and safety monitoring to partition software. \textsc{Jiao} fills this gap with robotics-specific parameter synchronization, IEC~61508-aligned integrity verification, and an independent hardware override~authority.

\noindent\textbf{Safety Communication Protocols.}
PROFIsafe~\cite{aakerberg2009exploring}, CIP Safety~\cite{vstohl2017cip}, and related protocols~\cite{peserico2021fsnetworks} implement the IEC~61508 black-channel principle over industrial fieldbuses. AUTOSAR~\cite{furst2016autosar} standardizes inter-ECU parameter management with configurations fixed at manufacturing. Our SCL adapts black-channel verification to intra-SoC shared memory, and our PSS addresses post-deployment parameter changes by non-expert integrators. Neither fieldbus protocols nor AUTOSAR were designed for this~scenario.

\noindent\textbf{Robotics Safety.}
Guiochet et~al.~\cite{guiochet2017robots} survey safety approaches for advanced robots and identify the gap between industrial standards and modern robotic flexibility. \textsc{Jiao} addresses this gap by making the isolation boundary transparent to integrators while preserving safety~guarantees.

\section{Conclusion}

We presented \textsc{Jiao}, a mixed-criticality robotics architecture that addresses expertise asymmetry through three integrated components, namely a Safe IO Cell, a Parameter Synchronization Service, and a Safety Communication Layer with IEC~61508-aligned verification. Evaluation demonstrates 84.5\% jitter reduction and nearly an order-of-magnitude improvement in tail timing error, eliminating all $|$jitter$|{>}50\,\mu$s excursions under concurrent perception~workloads.

\section*{Acknowledgment}

This work was supported in part by the National Natural Science Foundation of China (Grants 62141218 and 62232012), the Shanghai Key Laboratory of Scalable Computing and Systems, and OpenMind (Wuhu) Intelligent Robot Co.,~Ltd.

\balance
\bibliographystyle{IEEEtranS}
\bibliography{references}

\end{document}